\documentclass[final,5p,times, twocolumn]{elsarticle}

\usepackage{hyperref}
\usepackage{lineno}
\usepackage{comment}
\usepackage{amsmath}
\usepackage{float}
\usepackage{graphicx}
\usepackage{svg}
\usepackage{pdfpages} 
\modulolinenumbers[5]
\usepackage{pgfplots}
\usepackage{amssymb}
\usepackage{multirow}
\usepackage{ gensymb }
\usepackage{multicol}
\usepackage{makecell}
\usepackage{caption}
\usepackage{fixltx2e}

\makeatletter
\g@addto@macro{\UrlBreaks}{\UrlOrds}
\makeatother

\usepackage{subfig}
\usepackage{caption}
\journal{arXiv}

\begin{document}
\begin{frontmatter}

\title{ \huge Deep coastal sea elements forecasting using U-Net based models}

\author{Jesús García Fernández} 
\ead{j.garciafernandez@student.maastrichtuniversity.nl}

\author{Ismail Alaoui Abdellaoui}
\ead{i.alaouiabdellaoui@student.maastrichtuniversity.nl}

\author{Siamak Mehrkanoon \corref{cor1}}
\ead{siamak.mehrkanoon@maastrichtuniversity.nl}

\cortext[cor1]{Corresponding author}

\address{Department of Data Science and Knowledge Engineering, Maastricht University, The Netherlands}

\begin{abstract}

The supply and demand of energy is influenced by meteorological conditions. The relevance of accurate weather forecasts increases as the demand for renewable energy sources increases. The energy providers and policy makers require weather information to make informed choices and establish optimal plans according to the operational objectives. Due to the recent development of deep learning techniques applied to satellite imagery, weather forecasting that uses remote sensing data has also been the subject of major progress. The present paper investigates multiple steps ahead frame prediction for coastal sea elements in the Netherlands using U-Net based architectures. Hourly data from the Copernicus observation programme spanned over a period of 2 years has been used to train the models and make the forecasting, including seasonal predictions. We propose a variation of the U-Net architecture and further extend this novel model using residual connections, parallel convolutions and asymmetric convolutions in order to introduce three additional architectures. In particular, we show that the architecture equipped with parallel and asymmetric convolutions as well as skip connections outperforms the other three discussed models.

\end{abstract} 

\end{frontmatter}

\section{Introduction}

Renewable energy system has received increasing attention in the last years. In this context, the ability to accurately predict weather elements is crucial to the effective use of weather elements resources.
In particular, it has been shown that weather forecasting affects sectors like agriculture, forestry, transportation and healthcare among others, thus having a major impact on the global economy \cite{katz2005economic, da2021novel, moreno2021hybrid,liu2021hybrid, moreno2020multi}. More importantly, weather prediction can be used to save thousands of human lives by being able to forecast various types of natural disasters like tornadoes and flash floods \cite{henderson2020hazard,simmons2005wsr}. 

Classical approaches to perform weather forecasting heavily relied on the thermodynamics laws, Navier-Strokes equations, the statistical properties of the data, as well as the various properties of the atmosphere \cite{lorenc1986analysis, bauer2015quiet, glahn1985statistical, holtslag1990high}. This set of methods belongs to the Numerical Weather Prediction (NWP) approaches and generally require a large amount of compute resources since the processing is done on supercomputers \cite{saito2011next}. 
 
Furthermore, it has been shown that NWP based approaches might suffer from computational instability, mainly due to the initial conditions of the models \cite{zhongzhen1986problems}. Recent data-driven approaches on the other hand perform a simulation of an entire system in order to predict its next state. The main methodology used by these methods is the usage of historical data to perform the forecasting. Based on the  success provided by machine learning models (i.e. support vector machines, random forests, gaussian processes and neural networks) to forecast time series, these approaches have also been used for weather data \cite{kim2003financial,dudek2015short,girard2003gaussian,radhika2009atmospheric,rasouli2012daily,trebing2020smaat,trebing2020wind}. In particular, neural networks based models can use either shallow or deep architectures. 
As opposed to shallow networks that require domain knowledge and feature engineering, deep convolutional neural networks are less constrained by domain expertise. Indeed, these networks are capable of extracting the underlying complex patterns of the data by stacking of multiple nonlinear layers. 
Deep learning based models have already shown promising results in weather elements forecasting as well as several other domains such as biomedical signal analysis, healthcare, neuroscience and dynamical systems among others \cite{webb2018deep,mehrkanoon2018deep,mehrkanoon2019deep,abdellaoui2020deep,mehrkanoon2012approximate,mehrkanoon2015learning,mehrkanoon2014parameter,mehrkanoon2019cross,ismail2,Tomasz}. In particular, convolutional neural networks has shown to be successful for a wide range of tasks \cite{krizhevsky2012imagenet,mehrkanoon2019deep}. The U-Net architecture which consists of two main parts, i.e. the contracting and expanding, has shown to be efficient in many computer vision tasks. It is efficient at processing the input data at lower resolutions and restore it to its original resolution. In particular, U-Net based architectures have been successfully applied to medical data for diverse tasks such as segmentation, cell counting, and reconstruction \cite{ronneberger2015u, falk2019u, han2018framing}. 

\begin{figure*}[!htbp]
\vspace{-0.6cm}
    \centering
    \includegraphics[scale=0.05]{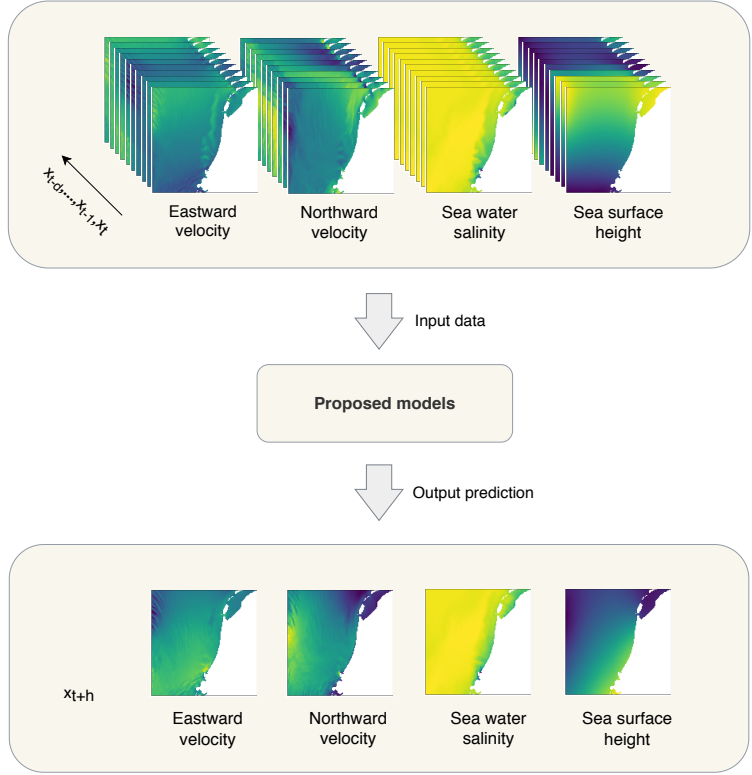}
    \caption{Prediction process overview. The time steps from \textit{t-d} to \textit{t} correspond to the lags of the models to generate the prediction of the time step \textit{t+h}, where \textit{d} is the number of lags and \textit{h} is the number of time steps ahead.}
    \label{fig:overview}
\end{figure*}

The main contribution of this paper is to extend U-Net based deep learning architectures to learn the underlying complex mapping between three-dimensional input data and two-dimensional output data. These architectures are then employed to perform multi-feature coastal weather forecasting as shown in Fig. \ref{fig:overview}. More precisely, we show that among the four novel proposed models, the one that uses a combination of asymmetric, parallel convolutions and skip connections inside each residual block outperforms the other models for the satellite imagery prediction of a coastal area of the Netherlands.

This paper is organized as follows. The literature review of satellite imagery for weather forecasting is discussed in section \ref{sec:related_work}. The proposed models are presented in section \ref{sec:proposedmodels}. Section \ref{sec:datadescription} introduces the used dataset. The experimental results, the corresponding discussion and finally the conclusion are given in sections \ref{sec:results} and \ref{sec:conclusion} respectively.

\section{Related Work}\label{sec:related_work}
Weather forecasting based on deep learning models has recently gain a lot of attention due to the rapid advancement of neural network techniques and availability of weather data \cite{mehrkanoon2019deep2}. The authors in \cite{zhou2019forecasting}, used a deep convolutional neural network to predict thunderstorms and heavy rains. The model was then compared against traditional machine learning models such as random forests and support vector machines. The authors in \cite{kim2017deeprain} incorporated multiple ConvLSTM layers to predict the precipitation rate using radar data. Moreover, multi-stream convolutional neural networks combines with a self-attention mechanism \cite{vaswani2017attention,ho2019axial} has been studied for precipitation forecasting \cite{sonderby2020metnet}.

In this paper we are interested in weather forecasting based on satellite imagery. Several approaches have been discussed in the literature for performing frame prediction from satellite data for weather forecasting using modern data-driven approaches. For instance, the authors in \cite{xiao2019spatiotemporal}, used satellite data in combination with deep learning techniques to perform sea surface temperature (SST) prediction in a subarea of the East China sea. The main type of layer used in this work was the ConvLSTM layer and was compared to three different models: support vector regression (SVR) model, a persistence model that was simulating a naive forecasting and a third model based on LSTM layers. It was shown that the ConvLSTM model outperformed the other models for $10$ days ahead prediction in a recursive fashion.

Similarly to the previous work, sea surface temperature forecasting is also performed using deep learning and remote sensing imagery from satellite data in \cite{zheng2020purely}.  
The methodology discussed in in \cite{zheng2020purely} is based on a multi-input convolutional neural networks that process the inputs using different spatial resolutions. The end goal of this work was to establish a relationship between sea surface temperature forecasting and tropical instability waves.

In \cite{shakya2020deep}, eight cyclone datasets are used for two main objectives: classifying whether the given image contains a storm or not as well as predicting the storm's location. In contrast with the previous works, this methodology is not end-to-end since multiple preprocessing steps are performed before training the deep learning model. In particular, multiple optical flow based techniques are used to perform temporal interpolation and the result of this processing is then fed to the deep learning models. The neural networks used are existing approaches that provide a fast inference time, namely YOLO \cite{redmon2016you} and RetinaNet \cite{lin2017focal}.

Another similar research work in \cite{rivolta2006artificial} performs precipitation nowcasting using artificial neural networks and satellite data. In this work, thermal infrared image prediction is first performed in order to get an estimation of the predicted precipitation. Hourly data is used and the neural network model is compared to other approaches such as linear interpolation, steady state methodology and persistence prediction.

Furthermore, future frame prediction is an active field of research in computer vision because of the several use cases where it can be applied such as anomaly detection, video prediction among others. In this context, within the field of deep learning, two main approaches are generally considered: autoencoders and generative adversarial neural networks \cite{liu2018future,patraucean2015spatio,hsieh2018learning,zhao2017spatio}.

\begin{figure*}[t]
\vspace{-0.5cm}
    \centering
    \includegraphics[scale=0.37]{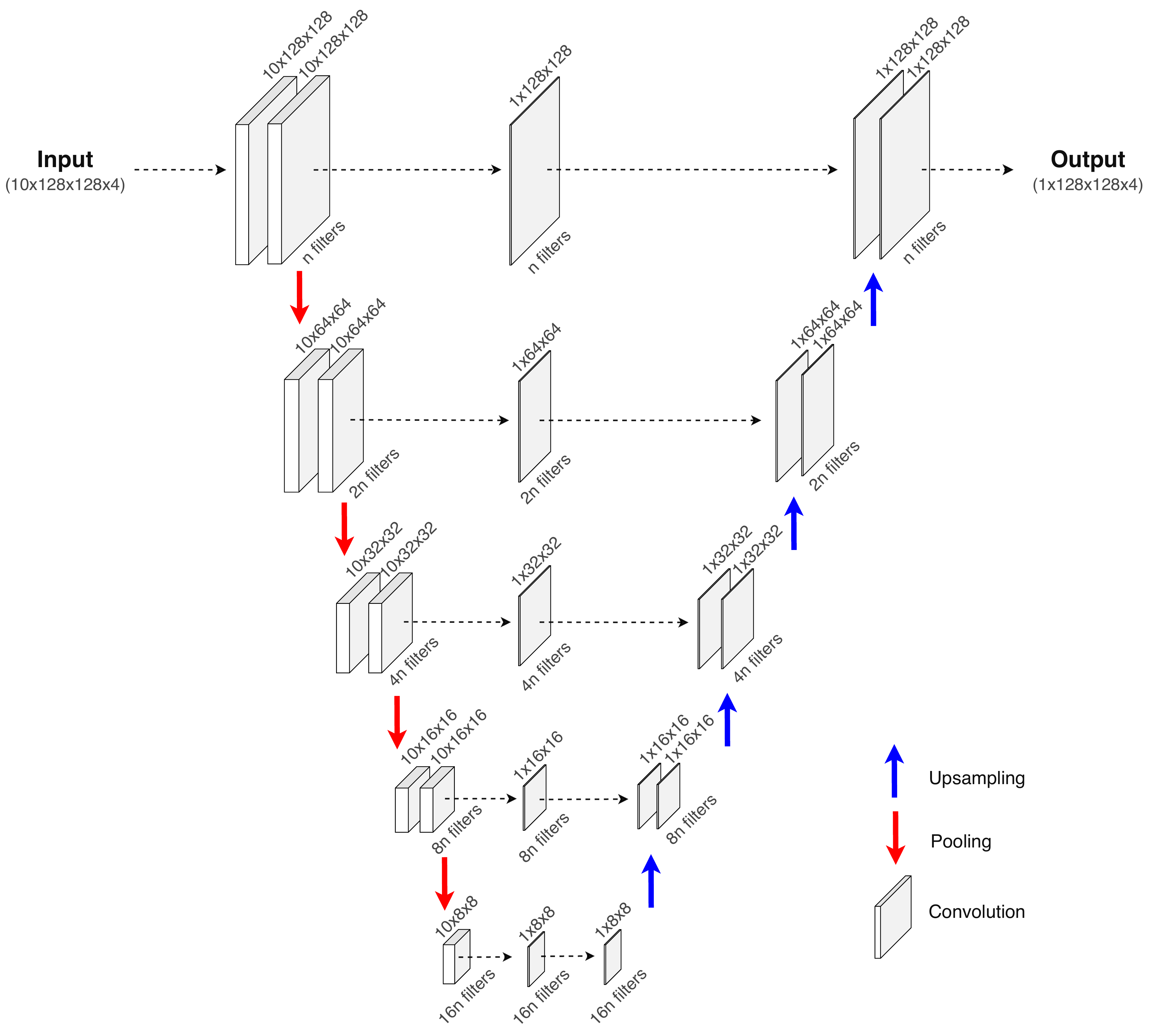}
    \caption{Architecture of 3DDR-UNet model. The annotations above the convolutions correspond to the output shape of those convolutions. Also, the number of filters is indicated below. We can appreciate that the first part reduces the dimensionality and then the second part upsamples the data to its original size (except for the temporal dimension). Between the reduction and expansion parts, intermediate convolutions reduce the temporal dimension (lags) from 10 to 1.}
    \label{fig:ArchModel1}
\end{figure*}

\section{Proposed models}\label{sec:proposedmodels}
We aim at proposing a model that accurately maps a set of input images to a set of output images. To this end, the UNet architecture \cite{ronneberger2015u} is used as the core model and is enriched by incorporating more advanced elements suitable for the task under study. The U-Net architecture is initially designed for medical image segmentation and has similar structure to that of an autoencoder. A first contracting part, where the features are extracted from the input image, is followed by an expanding part that performs classification on each pixel. 

In this paper, we propose an extended UNet architecture. Additionally modern enhancement techniques such as residual connections \cite{he2016deep}, inception modules \cite{szegedy2015going, szegedy2016inception} and asymmetric convolutions \cite{yang2019asymmetric} are taken into account when designing these models. The residual or skip connections have shown to improve the performance on deep networks by avoiding the vanishing of small gradients. On the other hand, inception modules apply convolutions with different kernels at the same level to capture features from larger and smaller areas in parallel. In the same way, asymmetric convolutions allow us to enlarge the network and thus its learning capacity, and at the same time the number of parameters is reduced.

In what follows, we propose four different models, each one being an extended version of the previous one. These models are 3DDR-UNet, Res-3DDR-UNet, InceptionRes-3DDR-UNet and AsymmInceptionRes-3DDR-UNet.

\subsection{3D Dimension Reducer UNet (3DDR-UNet)}

\begin{figure*}[!htbp]
\vspace{-0.5cm}
    \centering
    \includegraphics[scale=0.3]{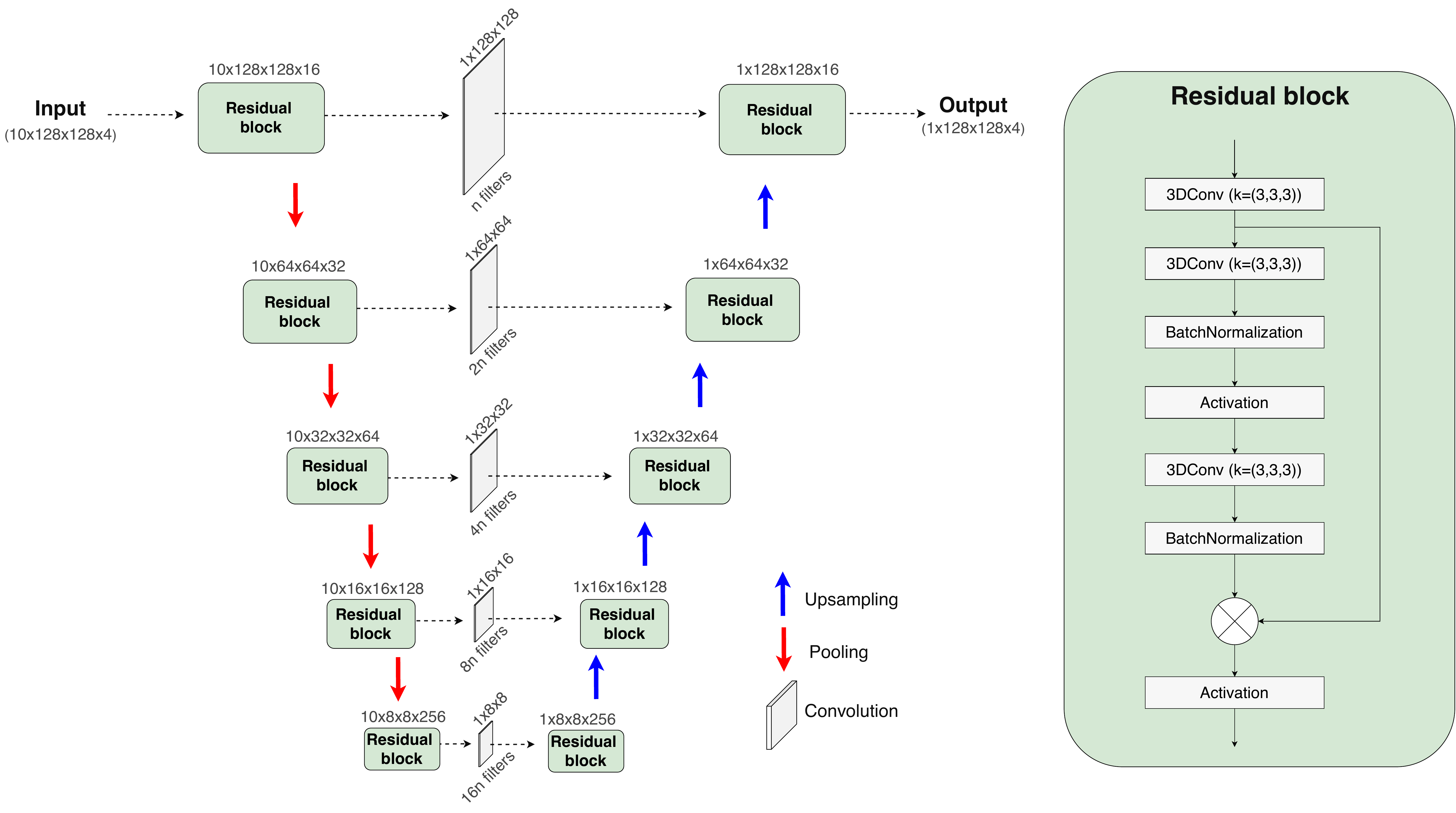}
    \caption{Architecture of Res-3DDR-UNet model. The annotations above the residual blocks correspond with the output shape of such blocks. Similarly to 3DDR-UNet, the first part reduces the dimensionality, and then the second part upsamples the data to its original size (except for the temporal dimension). Between the reduction and expansion parts, intermediate convolutions reduce the temporal dimension (lags) from 10 to 1.}
    \label{fig:ArchModel2}
\end{figure*}

This section introduces 3D Dimension Reducer UNet (3DDR-UNet) which is based on UNet core architecture \cite{ronneberger2015u}.
The classical UNet model, is a fully convolutional neural networks containing two parts: A contraction part or encoder and an expansion part or decoder. The first part is composed of stacked convolutions and pooling operations to extract features and capture the context in the input. The second symmetric part combines the features extracted in the contraction part with an upsampled output. In this way, the network expands the data to its original size and projects the learned features onto the pixel space to perform an accurate classification of them. 

Here, we propose the 3DDR-UNet architecture, which manipulates 3-dimensional data in the encoder and 2-dimensional data in the decoder. This configuration allows the network to capture spatial and temporal dependencies from a stack of 2-dimensional images in the contracting part. Then the first input dimension (time dimension) is reduced from n (number of time-steps or lags, which is 10 in our case) to 1 in the middle of the network before the data moves towards the expanding part. Those extracted and combined features are later used to reconstruct one single image in the decoder. The reduction in the first input dimension is carried out by convolutions with kernel size $n\times1\times1$ ($10\times1\times1$ in our case) and valid padding. The output of these operations is a weighted average of different time-steps (i.e. lags) in the input. Essentially, this architecture extracts features on the encoder part and averages them in a weighted fashion over the first dimension before it is fed into the decoder part. The number of convolutional filters grows exponentially after each pooling from n to 16n in the encoder part, and shrinks again to n in the decoder part, with a kernel size of $3\times3\times3$. The size of both the pooling and the upsampling operations is set to $1\times2\times2$. In this way, the temporal dimension of the data remains unchanged during the pooling and upsampling, while the spatial dimension of the data (second and third dimension) is reduced and later upsampled. 

Given the nature of the task, we train the network to perform a regression of every pixel. Traditionally UNet is used for segmentation tasks. Here, as opposed to the segmentation tasks, in which each pixel belongs to a class, we first normalize both input and output data. Then the Mean Squared Error (MSE) metric is used during the training to minimize the difference between the predicted value of each pixel and the ground truth value. The architecture 3DDR-UNet of the network is shown in Fig. \ref{fig:ArchModel1}.

\subsection{Residual 3D Dimension Reducer UNet (Res-3DDR-UNet)}

\begin{figure*}[!htbp]
\vspace{-0.5cm}
    \centering
    \includegraphics[scale=0.27]{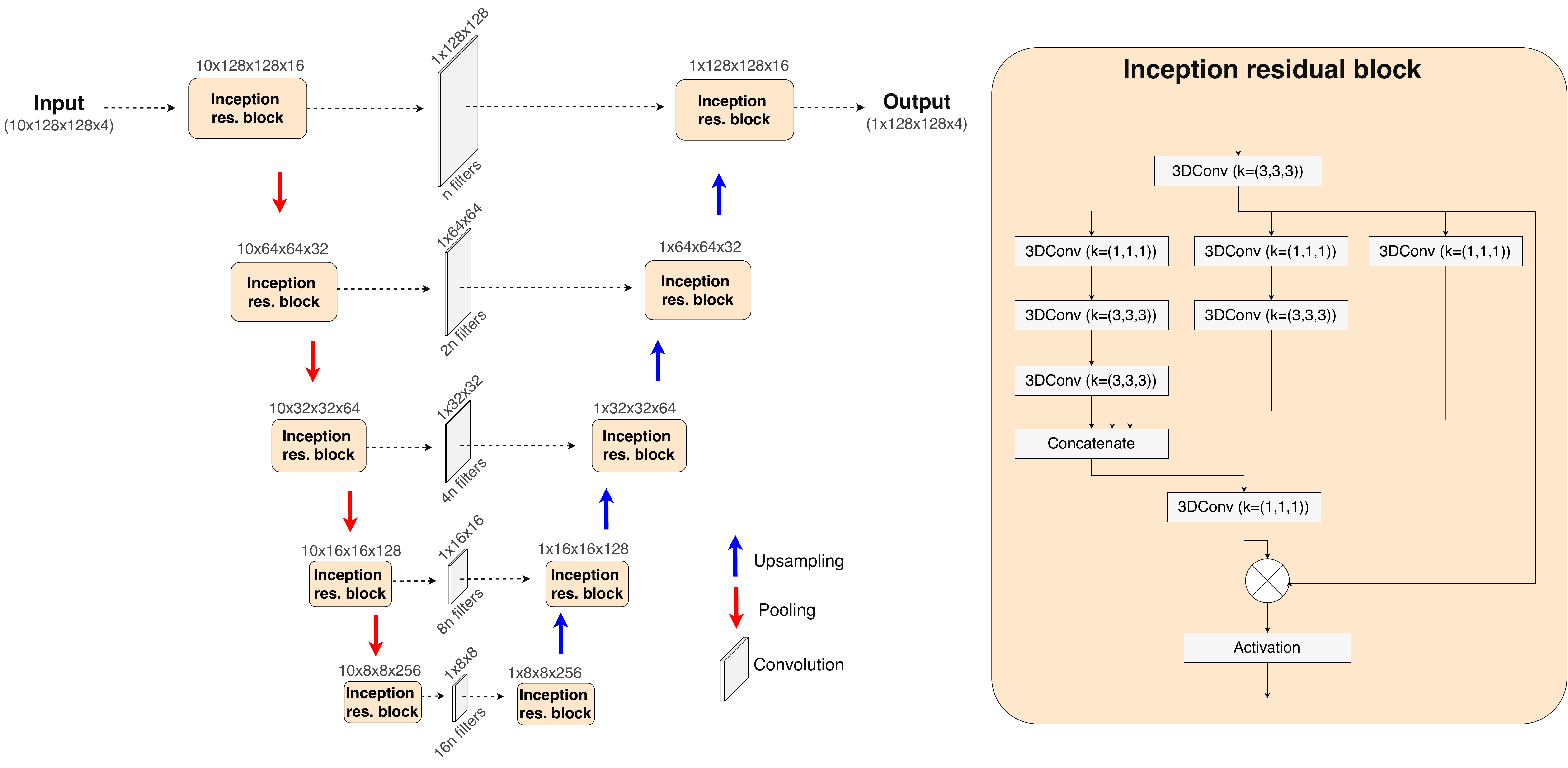}
    \caption{Architecture of InceptionRes-3DDR-UNet model. The annotations above the inception residual blocks correspond to the output shape of such blocks. Similarly to the previous models, the first part reduces the dimensionality, and then the second part upsamples the data to its original size (except for the temporal dimension). Between the reduction and expansion parts, intermediate convolutions reduce the temporal dimension (lags) from 10 to 1.}
    \label{fig:ArchModel3}
\end{figure*}

The second proposed model, Residual 3D Dimension Reducer UNet (Res-3DDR-UNet), is an extension of 3DDR-UNet model introduced previously. In order to augment its learning capacity, we scale up the model by adding a generous number of convolutional operations. We use three convolutional layers and a skip connection around the first layer and the final activation to avoid the vanishing of gradients. All these operations form a residual block.

Further, the outputs of the last convolutions in the block are normalized making use of a batch normalization layer. This normalization also aims at increasing the robustness of the network and alleviate the vanishing gradient problem.

Following the lines of \cite{he2016deep}, we skip two convolutional layers in each block which enhances the performance as well as faster training. As a result of these changes, the number of trainable parameters grows by 50\% compare to the 3DDR-UNet. It should be noted that the kernel size of the convolutions, pooling and upsampling as well as the loss function were similar to the ones used in 3DDR-UNet. The architecture of the Res-3DDR-UNet is depicted in Fig. \ref{fig:ArchModel2}.

\subsection{Inception Residual 3D Dimension Reducer UNet (InceptionRes-3DDR-UNet)}

Motivated by the effectiveness of inception modules in CNN classifiers \cite{szegedy2015going, szegedy2016rethinking}, we include similar modules in our residual blocks. Within this module, the data stream is splitted into parallel convolutions with different kernel sizes. Later the branches are concatenated again.
This structure is motivated by the ability to extract various features through the usage of multiple kernel sizes applied to the same data. After these parallel operations, the features are concatenated and combined with a $1\times1\times1$ convolution, which is equivalent to a weighted average. Hence, the network learns to favor over time the branches with the most suitable kernels. Essentially this module allows the network to employ different kernels for the task, and give more importance to the most relevant ones.

Here in particular, we use three parallel branches with $1\times1\times1$, $3\times3\times3$ and $5\times5\times5$ kernels. As suggested in \cite{szegedy2016rethinking}, we approximate the $5\times5\times5$ convolution by two sequential $3\times3\times3$ convolutions, leading to a reduction in the computational cost. In addition, a convolution of $1\times1\times1$ is included at the beginning of each branch to reduce the dimensionality of the data and thus reducing the computational cost. Furthermore, inspired by the performance of \cite{szegedy2016inception}, we kept the residual connection that skips the parallel branches. The number of convolutional filters, kernel sizes of pooling and upsampling as well as the loss function are the same as those of the previous models. The architecture of InceptionRes-3DDR-UNet is shown in Fig. \ref{fig:ArchModel3}.

\subsection{Asymmetric Inception Residual 3D Dimension Reducer UNet\newline (AsymmInceptionRes-3DDR-UNet)}

\begin{figure}[!htbp]
    \centering
    \includegraphics[scale=0.2]{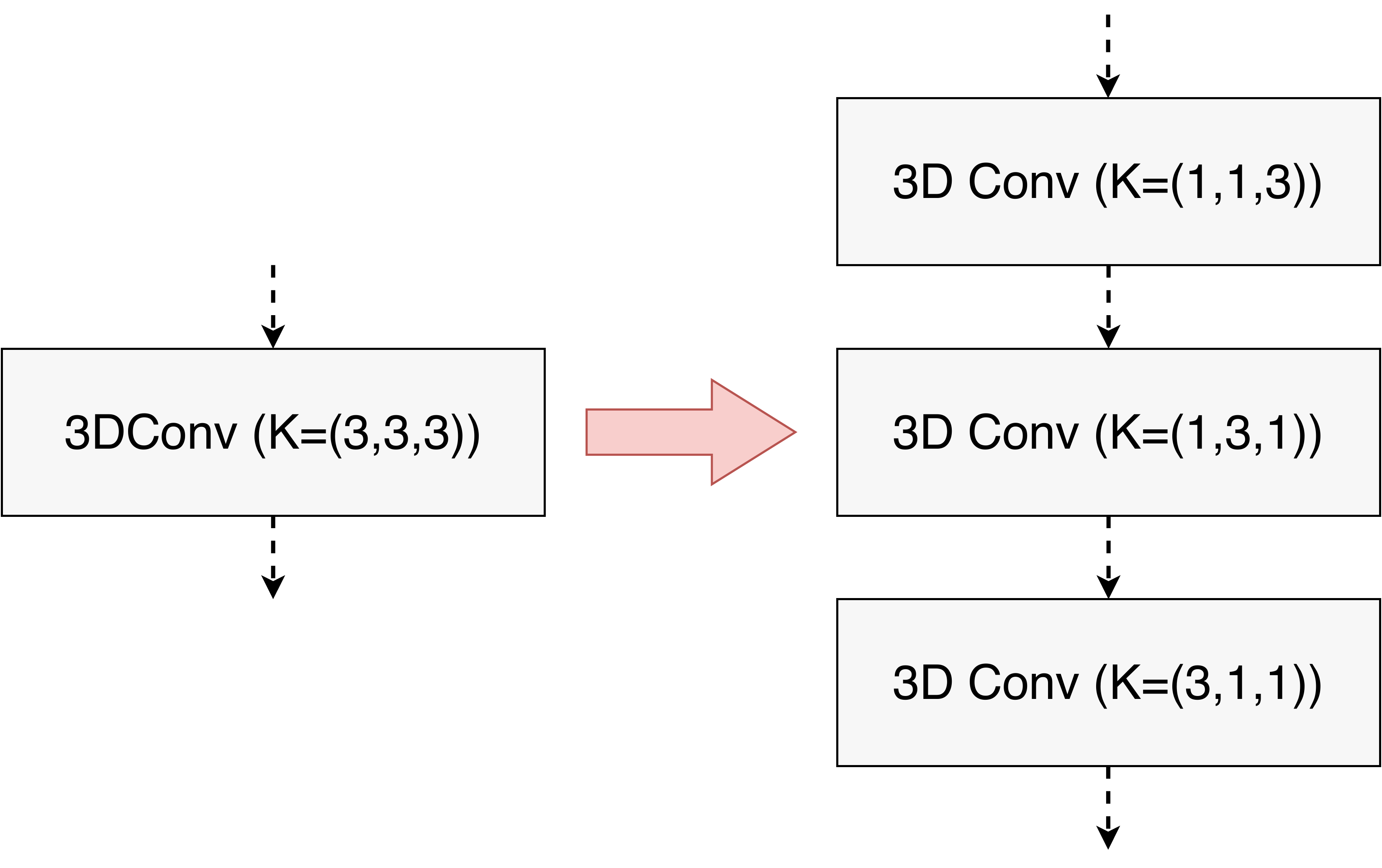}
    \caption{Example of kernel decomposition in the asymmetric convolution operation.}
    \label{fig:AsymmConv}
\end{figure}

\begin{figure*}[!htbp]
    \centering
    \includegraphics[scale=0.25]{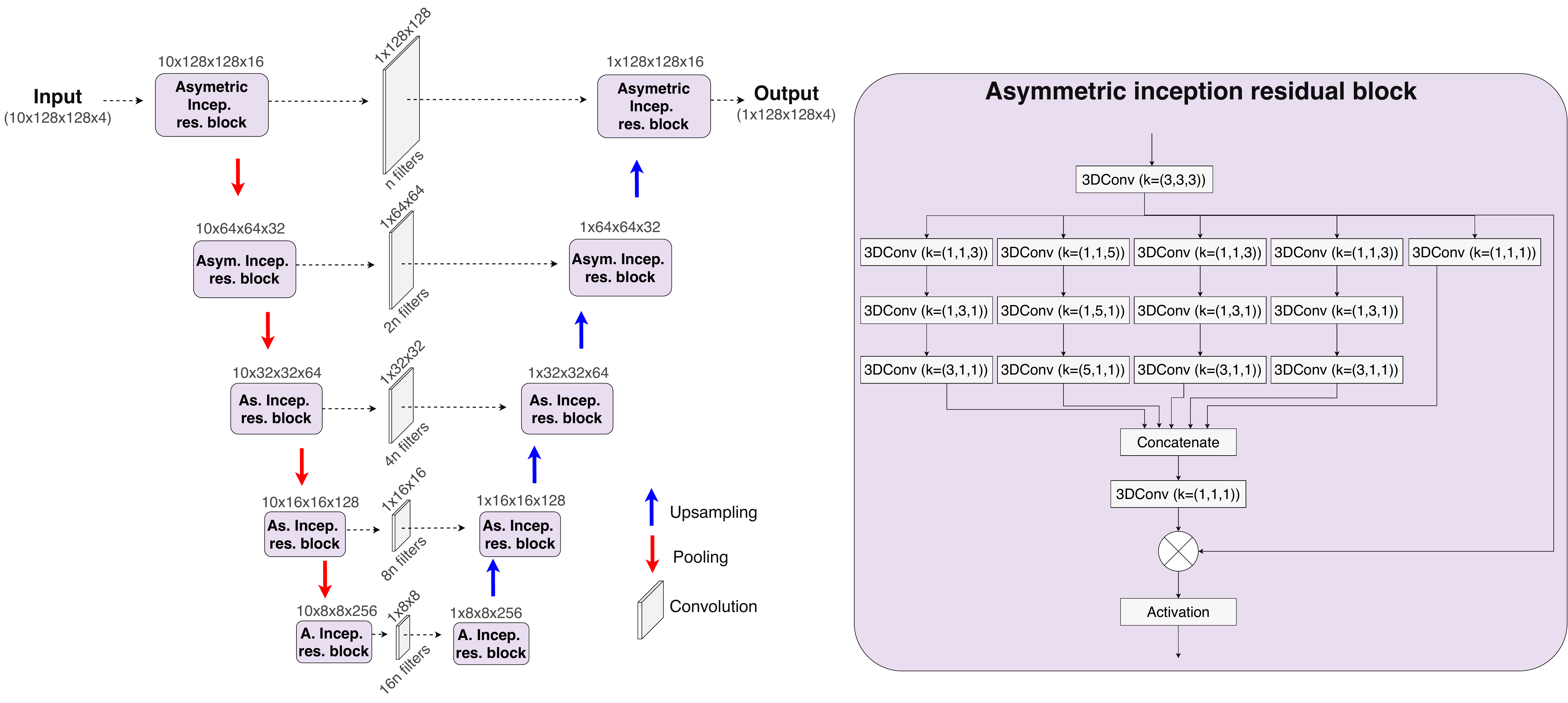}
    \caption{Architecture of AsymmInceptionRes-3DDR-UNet model. The annotations above the asymmetric inception residual blocks correspond to the output shape of such blocks. Similarly to the previous models, the first part reduces the dimensionality, and then the second part upsamples the data to its original size (except for the temporal dimension). Between the reduction and expansion parts, intermediate convolutions reduce the temporal dimension (lags) from 10 to 1.}
    \label{fig:ArchModel4}
\end{figure*}

Driven by the need to reduce the parameters in InceptionRes-3DDR-UNet, we introduce a lighter, yet more effective model. Here we use asymmetric convolutions \cite{yang2019asymmetric} to lower the complexity of the parallel convolutions. As shown in Fig. \ref{fig:AsymmConv}, each kernel is decomposed into three simpler ones and applied consecutively.
The resulting combination of operations is an approximation of the original operation with considerably fewer parameters (see \cite{yang2019asymmetric} for more details). In consequence of such reduction of parameters, we can afford to remove the $1\times1\times1$ convolution at the beginning of each branch within the asymmetric inception residual block, whose purpose is to reduce the complexity of the data and thus making the model lighter. Furthermore, we have added two more parallel branches in each block to make the model comparable to the previous ones in terms of parameters. The number of convolutional filters, the kernel size of pooling and upsampling as well as the loss function are the same as in the previous described models. The architecture of AsymmInceptionRes-3DDR-UNet is depicted in Fig.  \ref{fig:ArchModel4}.

\section{Data Description}\label{sec:datadescription}
The data used in this paper consists of satellite images. It is provided by Copernicus \footnote{\url{https://marine.copernicus.eu/}},
the observation program led by the European Commission and the European Space Agency (ESA). Specifically, it is part of the dataset "Atlantic - European North West Shelf - Ocean Physics Analysis and Forecast" \cite{copernicusDataset}, covering a geographical area with longitude from E 002\degree000 to E 006\degree000 and latitude from N 51\degree600 to N 53\degree400. The spatial resolution is approximately 1.5 km, so every pixel represents a region of size $1.5\times1.5$ km. We chose such a geographical area since it covers both the land and sea of the Netherlands. Furthermore, the selected observations start on 01/03/2017 and end on 13/02/2019, with an hourly temporal resolution. The dataset consists of four weather variables, i.e. Eastward current velocity (EastCUR), Northward current velocity (NorthCUR), Seawater salinity (SAL) and Sea surface height (SSH).

\begin{figure}[!htbp]
    \centering
    \includegraphics[scale=0.45]{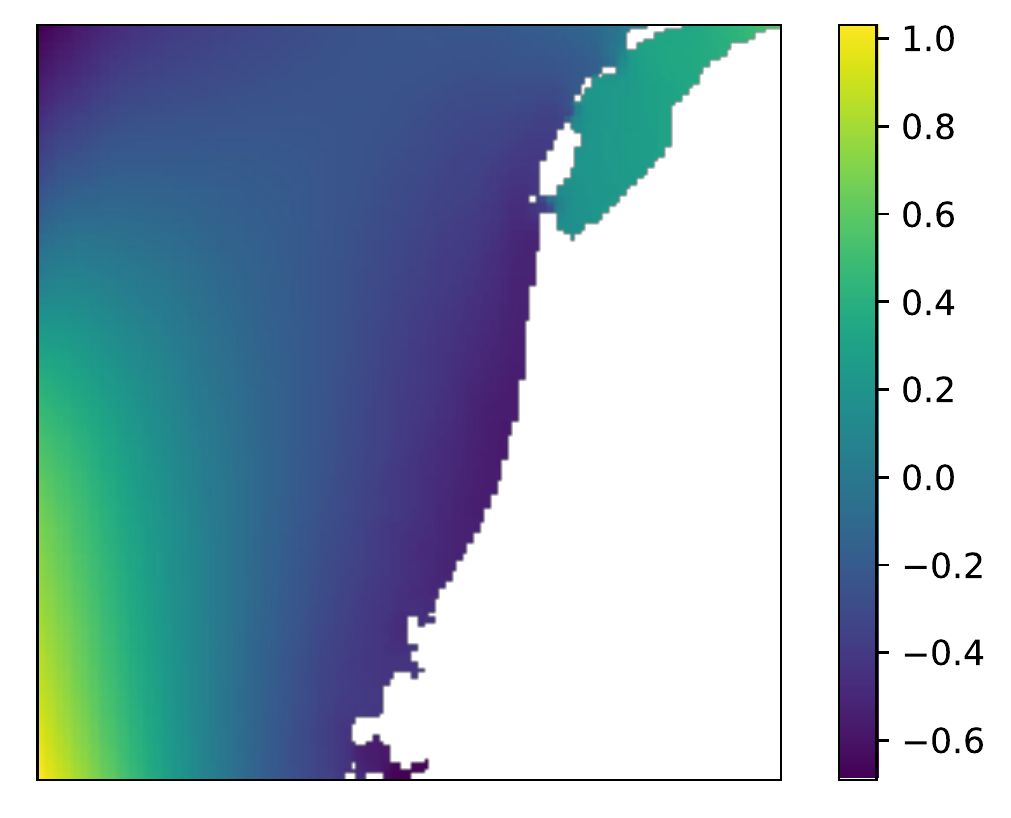}
    \caption{Example of the sea surface height in meters of the studied region.}
    \label{fig:ExampleSeaLevel}
\end{figure}

Therefore, each time-step of each variable is represented by $135\times135$ image (see Fig. \ref{fig:ExampleSeaLevel}). For reproducibility purposes, the code as well as the dataset are available on Github \footnote{\url{https://github.com/jesusgf96/Sea-Elements-Prediction-UNet-Based-Models}}. More details on the included variables, can be found in the official documentation\footnote{\url{https://resources.marine.copernicus.eu/documents/PUM/CMEMS-NWS-PUM-004-013.pdf}}.

\section{Experimental Results}\label{sec:results}
\subsection{Data Preprocessing}

As the data contains sea elements, the pixels in the image that represent the ground should not be taken into account by the network.
In practice, the pixels corresponding to the land are masked initially. Hence, we apply a MinMax scaling with a boundary of 0.1 and 1 to the pixels representing the sea elements as follows:

\begin{equation}
    x = 0.1+\frac{((x-x_{min})*(1-0.1))}{x_{max}-x_{min}}.
\end{equation}

Then the pixels representing the ground are assigned a zero value. In this way, the pixels that belong to the ground are invisible to the models because of the ReLu activation function used within them. Moreover, we crop the images seven pixels from the right and the bottom sides, resulting in a  $128\times128$ shape, which is suitable for the subsequent convolutional and pooling  operations. The data is arranged in such a way that the resulting object is a four-dimensional array $\mathcal{T} \in \mathbb{R}^{L \times H \times W \times V}$, where $L$ is the number of timesteps, which makes up the time dimension. $H$ and $W$ refer to the size of the image and form the spatial dimensions. The last element $V$ corresponds to the sea variables.

\subsection{Experimental Setup}
For all the models, we use all the variables as input, and we perform a prediction of the same variables. The number of convolutional filters is chosen in such a way that all models contain comparable total number of trainable parameters.

In our experiments, the number of lags is set to 10 as empirically it was found to yield better performance compared to other tested lag values. Therefore, the model receives ten hours of information to predict one single time-step, which translate into having an input with shape (10, 128, 128, 4) and output with shape (1, 128, 128, 4). In addition, to test the predictive ability of the models as well as their robustness, four different experiments are carried out. It consists of performing 12, 24, 48, and 72 hours ahead predictions for all the variables. The models are trained with data spanned over a year, from 01/03/2017 to 01/03/2018, in total 8760 hours training data. The validation data is composed of 2016 hours, and represents 504 hours from each season (spring, summer, autumn and winter). This validation data corresponds to a period of time after the training data. Similarly, the test data is composed of another 2016 hours after the training data. The specific days used in both the validation and test can be found in Table \ref{tab:timeStepsValTest}.

\begin{table}[!htbp]
    \centering
    \renewcommand{\arraystretch}{1.5}
    \caption{Time-steps used for validation and test.}
    \label{tab:timeStepsValTest}
    \resizebox{\columnwidth}{!}{
    \begin{tabular}{c c c c c c}
    \Xhline{3\arrayrulewidth}
    \multirow{2}{*}{}&\multirow{2}{*}{} & \multirow{2}{*}{\makebox{\textbf{Spring}}} & \multirow{2}{*}{\makebox{\textbf{Summer}}} & \multirow{2}{*}{\makebox{\textbf{Autumn}}}& \multirow{2}{*}{\makebox{\textbf{Winter}}}\\
    & & & & \\\Xhline{3\arrayrulewidth}
        \multirow{2}{*}{\textbf{Validation}} & \textbf{From} & 01/04/2018 & 01/07/2018 & 01/10/2018 & 01/01/2019 \\
        & \textbf{To} & 22/04/2018 & 22/07/2018 & 22/10/2018 & 22/01/2019 \\ \hline
        \multirow{2}{*}{\textbf{Test}} & \textbf{From} & 23/04/2018 & 23/07/2018 & 23/10/2018 & 23/01/2019 \\
        & \textbf{To} & 13/05/2018 & 12/08/2018 & 13/11/2018 & 13/02/2019\\ 
        \Xhline{3\arrayrulewidth}
    \end{tabular}
    }
\end{table}

\subsection{Training}
The same training setup is used in all the models. As mentioned previously in section \ref{sec:proposedmodels}, the Mean Squared Error (MSE) is used as the loss function to minimize the differences between the predicted and the ground truth image. Adam optimization method \cite{kingma2014adam} is applied to optimize the loss function. The batch size and the dropout rate are set to 16 and 0.5 respectively. We also implemented a checkpoint callback that monitors the validation loss and we let the models to be trained for 100 epochs in each of the experiments. The best results are then saved based on the performance of the models on the validation data.

\subsection{Results and discussion}
\begin{figure}[!h]
    \centering
        \includegraphics[scale=0.28]{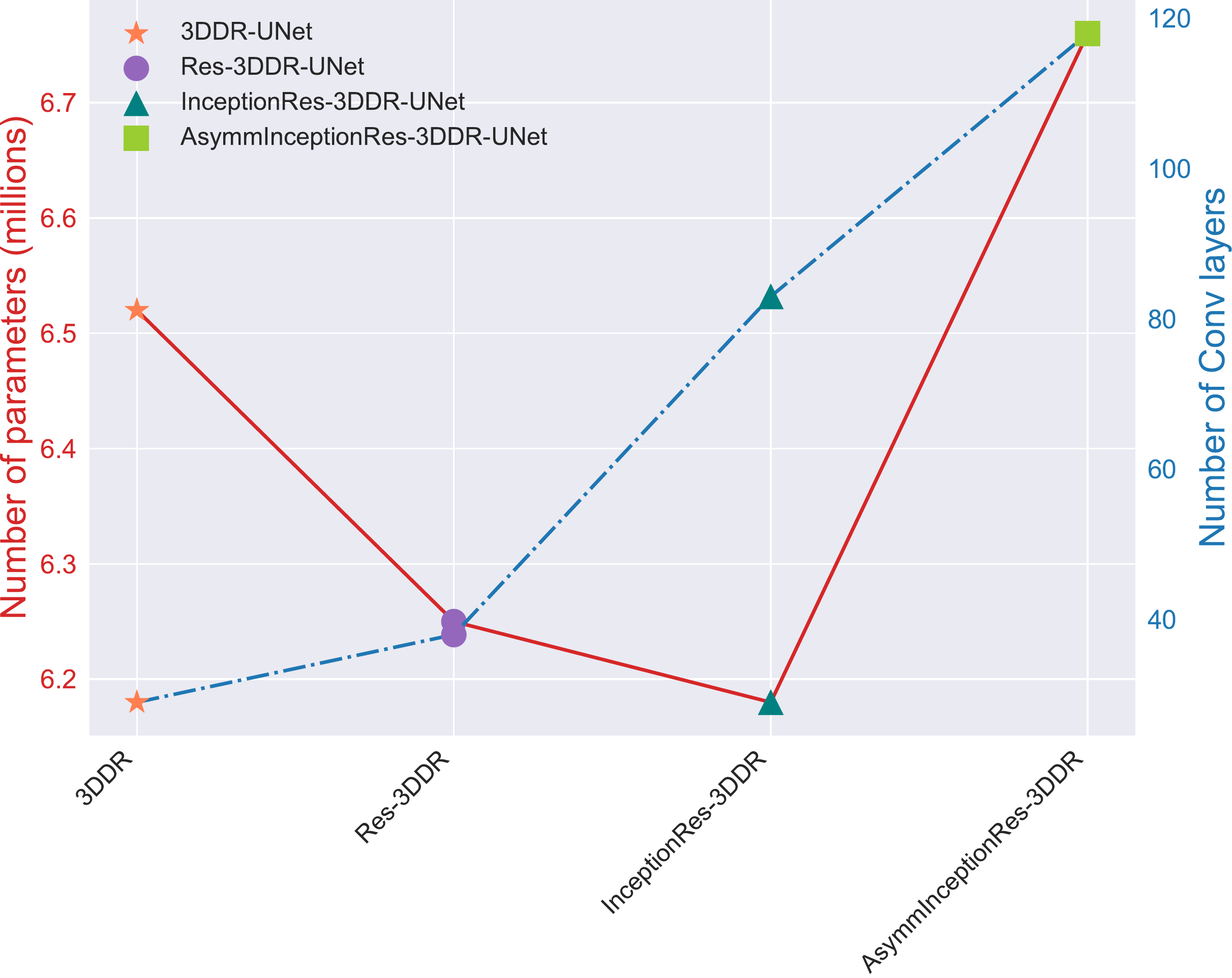}
    \caption{Comparison of the number of parameters and convolutional layers between models.}
    \label{fig:numberParams}
\end{figure}

\begin{figure*}[!htbp]
\centering
    \subfloat[]{\includegraphics[scale=0.35]{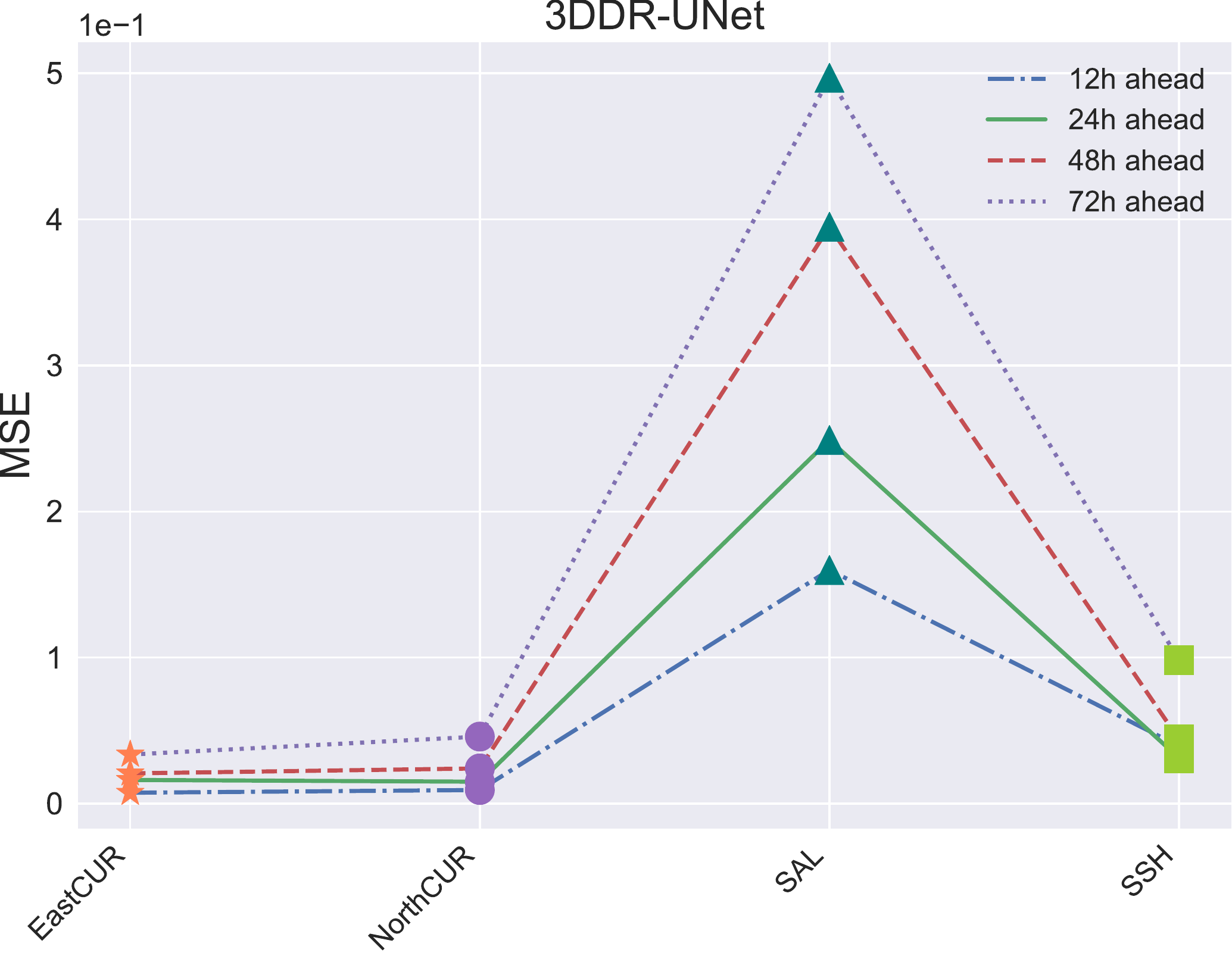}}
    \hspace{0.4cm}
    \subfloat[]{\includegraphics[scale=0.35]{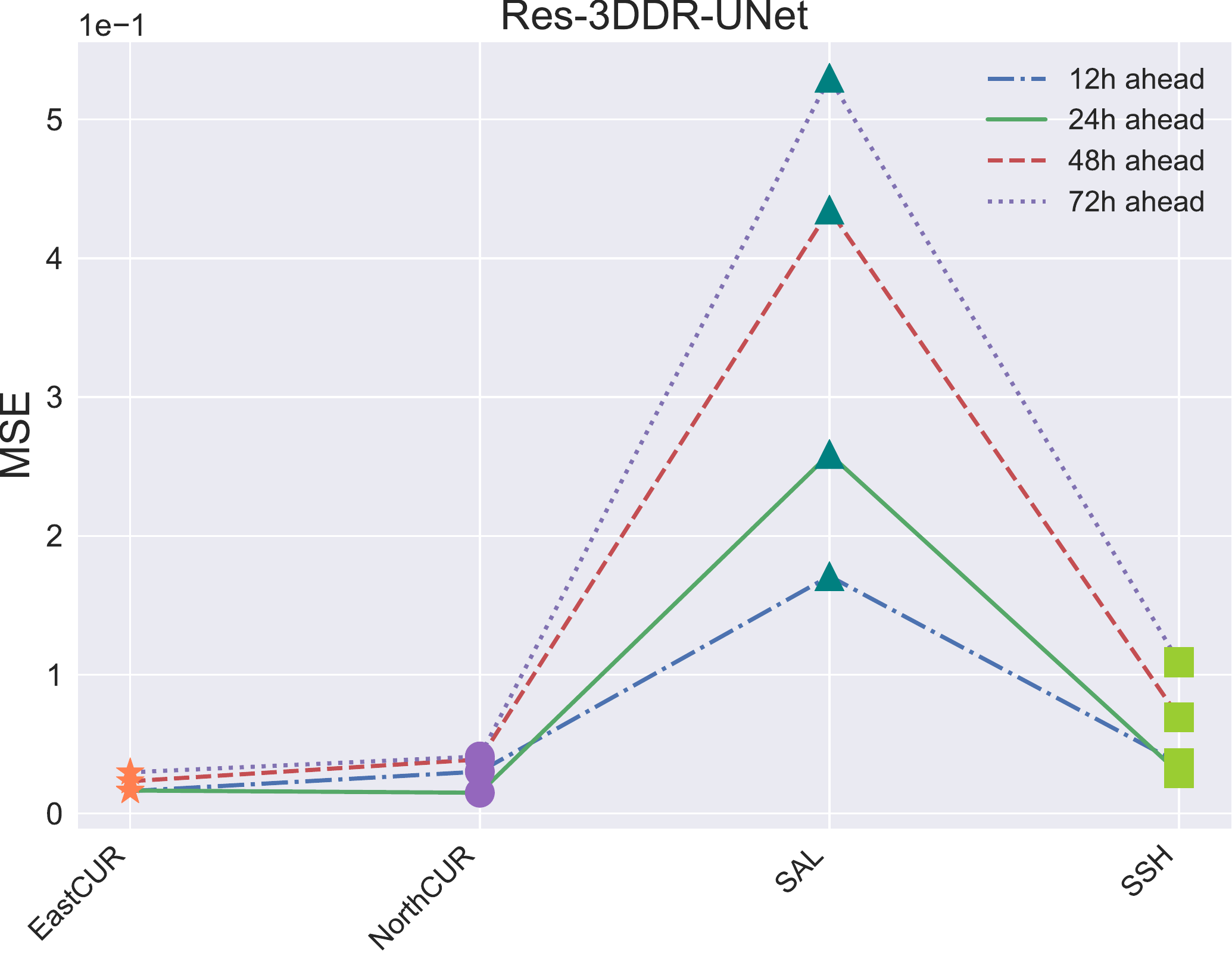}}
    \\\vskip 0.5pt plus 0.25fil
    \subfloat[]{\includegraphics[scale=0.35]{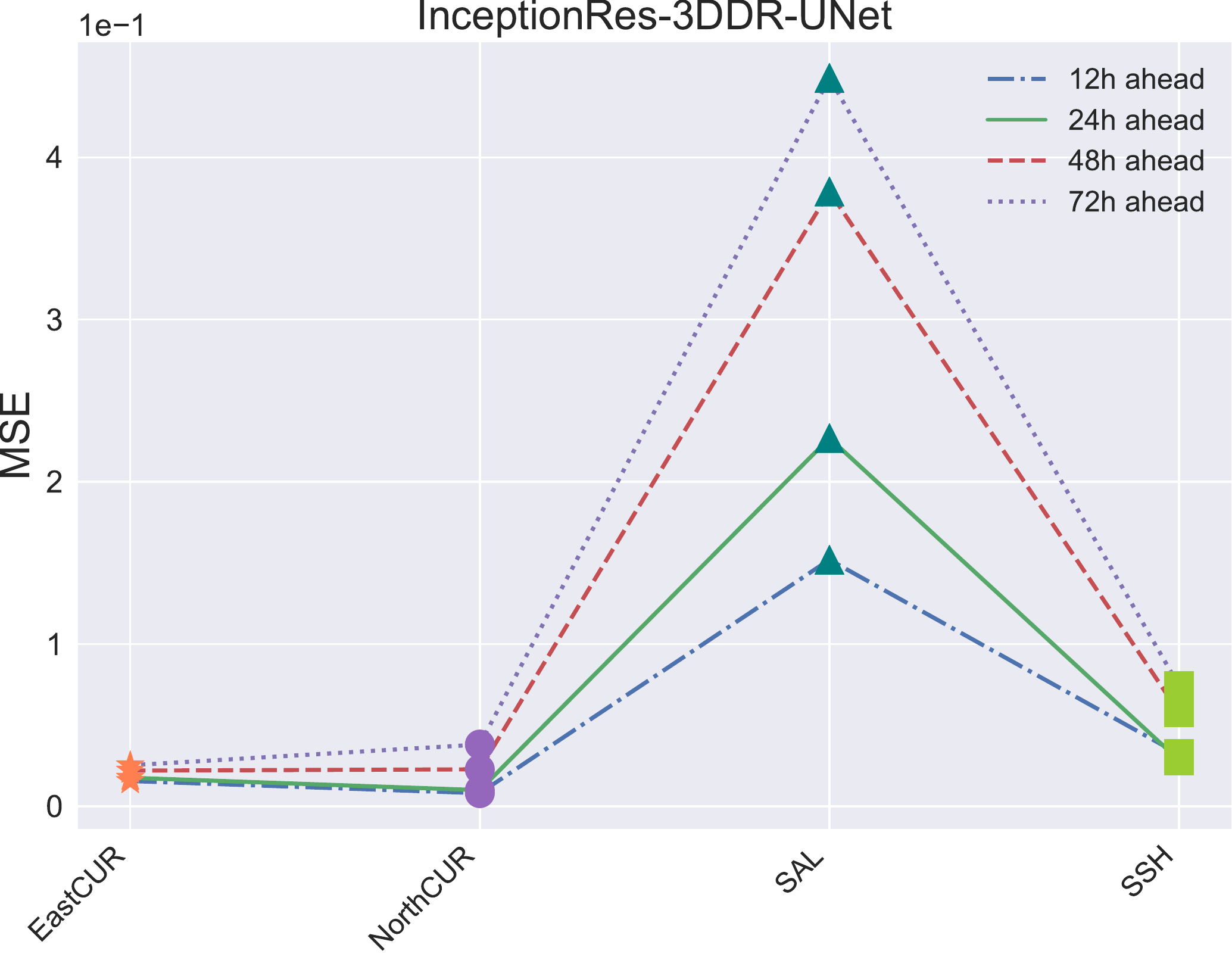}}
    \hspace{0.4cm}
    \subfloat[]{\includegraphics[scale=0.35]{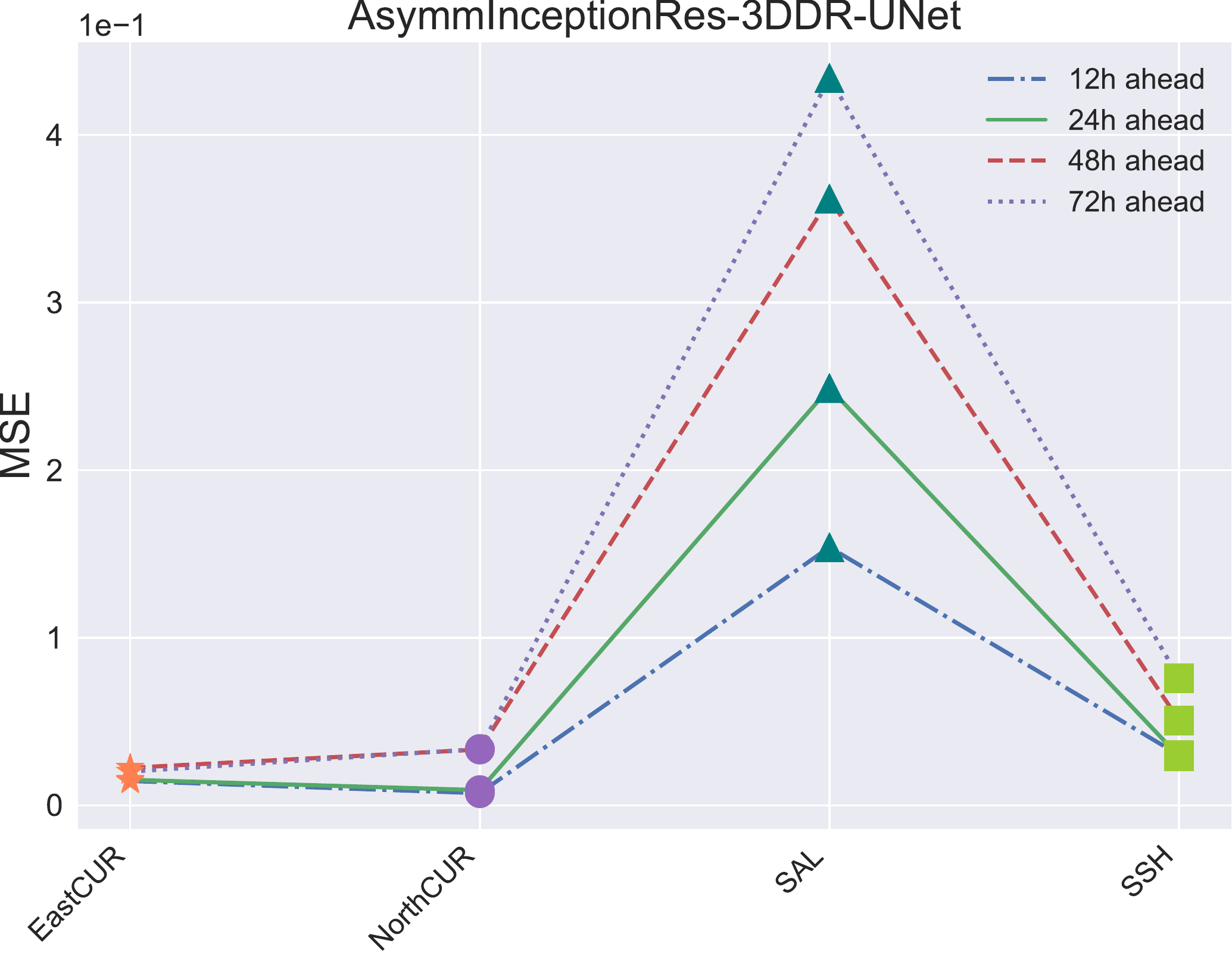}}
    \\\vskip 0.5pt plus 0.25fil
    \caption{MSE of individual sea elements for different number of hours ahead. The predictions are performed with the (a) 3DDR-UNet, (b) Res-3DDR-UNet, (c) InceptionRes-3DDR-UNet, and (d)AsymmInceptionRes-3DDR-UNet trained models.}
    \label{fig:mseVariables}
\end{figure*}

\begin{figure*}[!htbp]
\centering
    \subfloat[]{\includegraphics[scale=0.35]{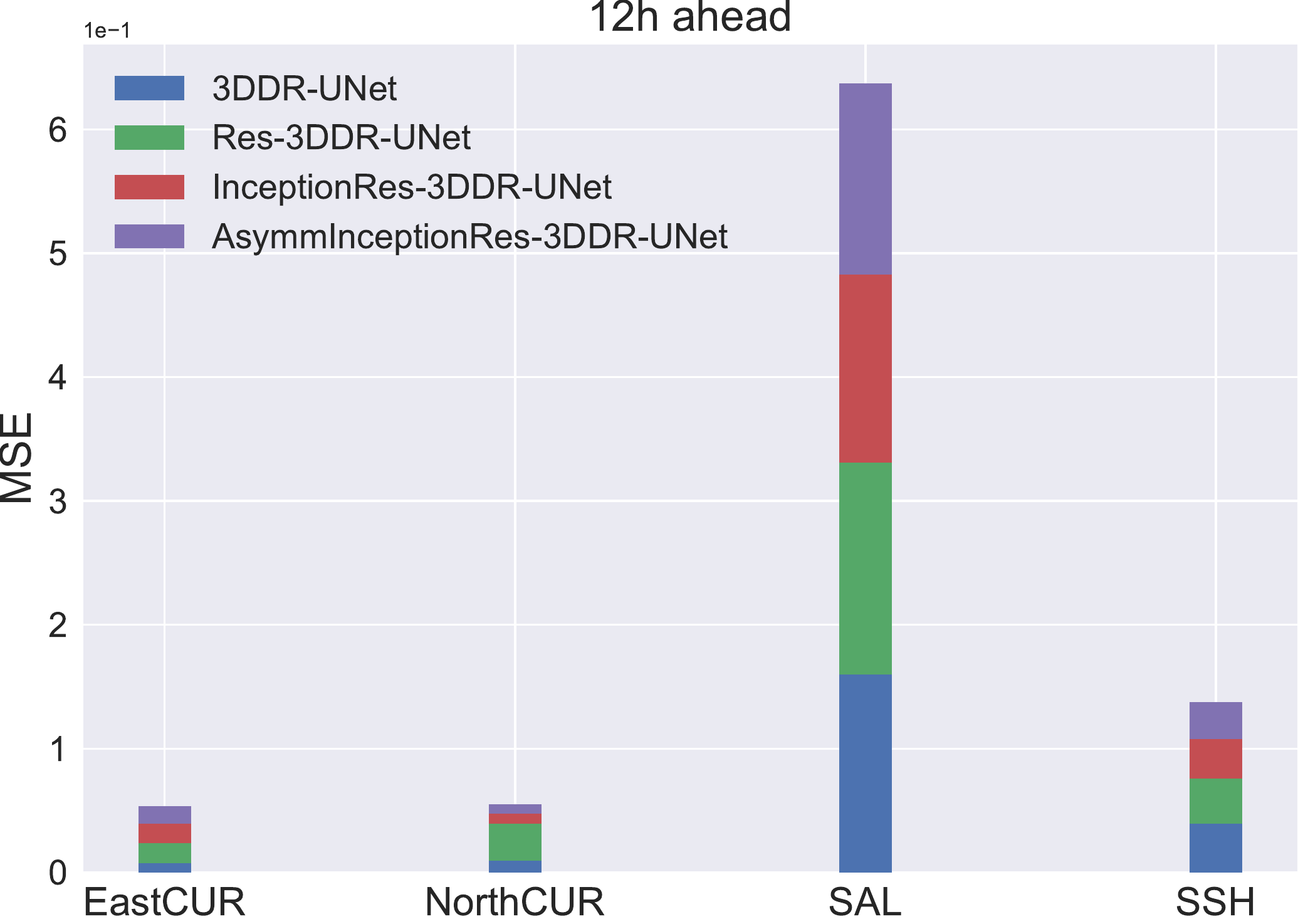}}
    \hspace{0.4cm}
    \subfloat[]{\includegraphics[scale=0.35]{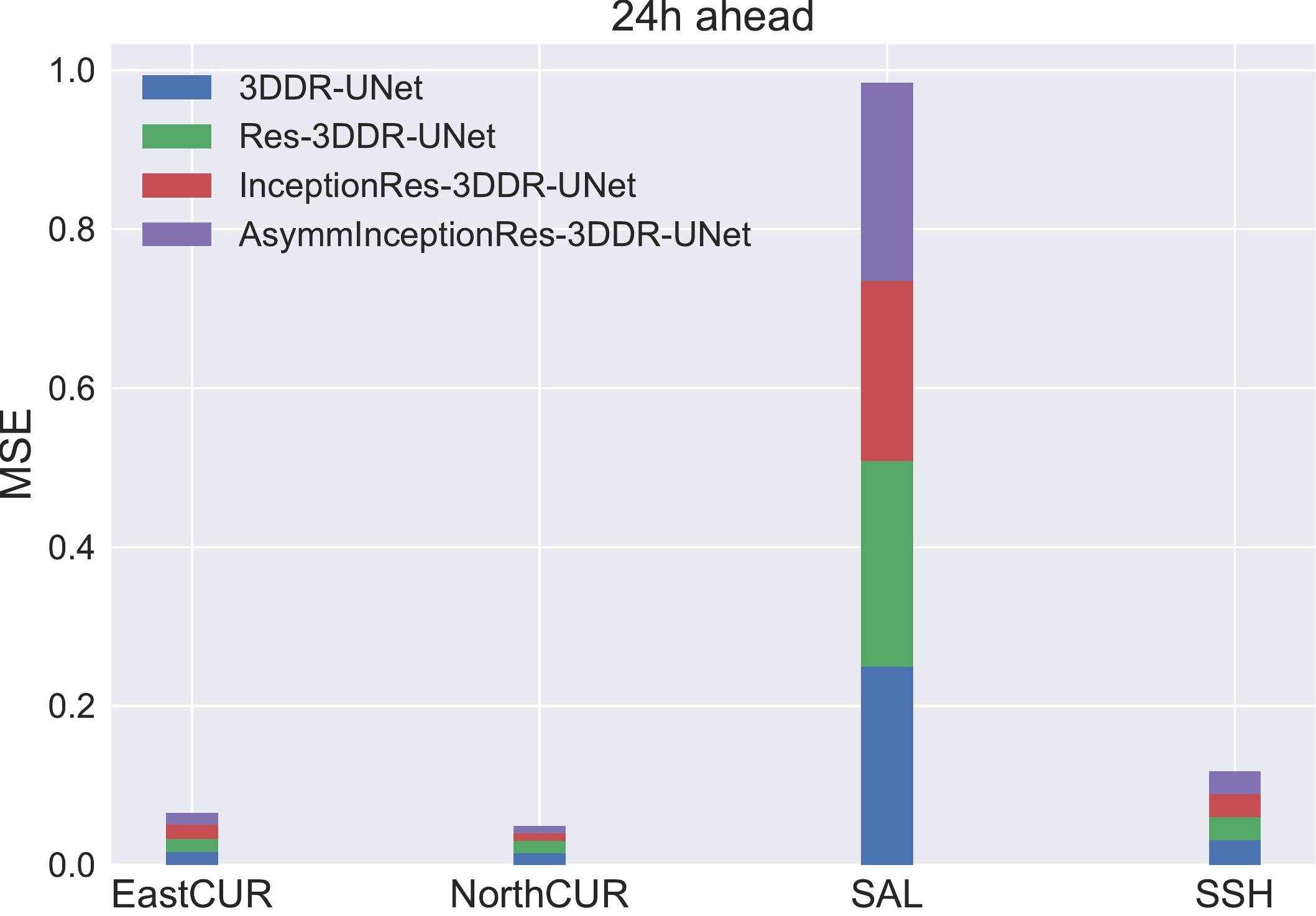}}
    \\\vskip 0.5pt plus 0.25fil
    \subfloat[]{\includegraphics[scale=0.35]{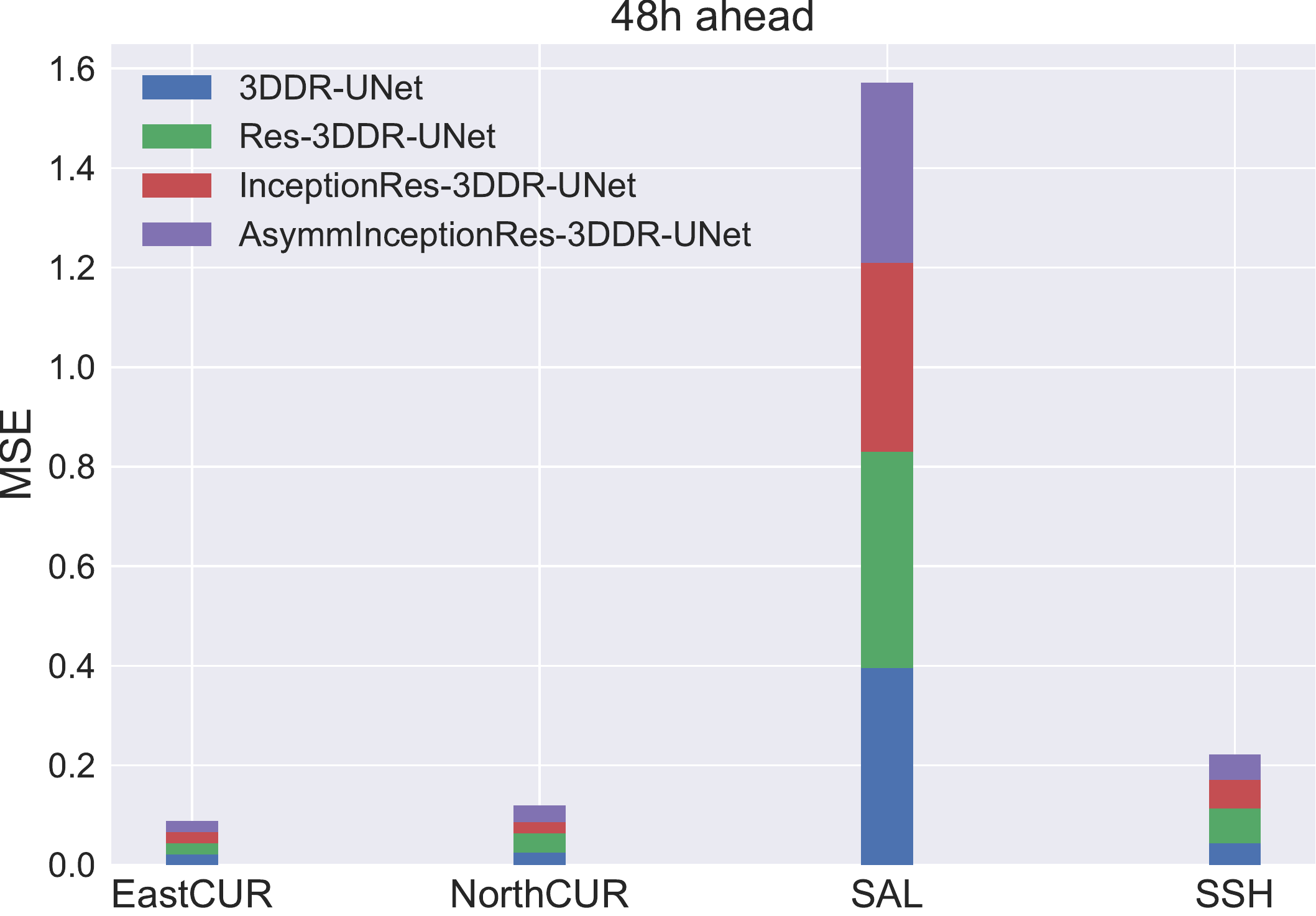}}
    \hspace{0.4cm}
    \subfloat[]{\includegraphics[scale=0.35]{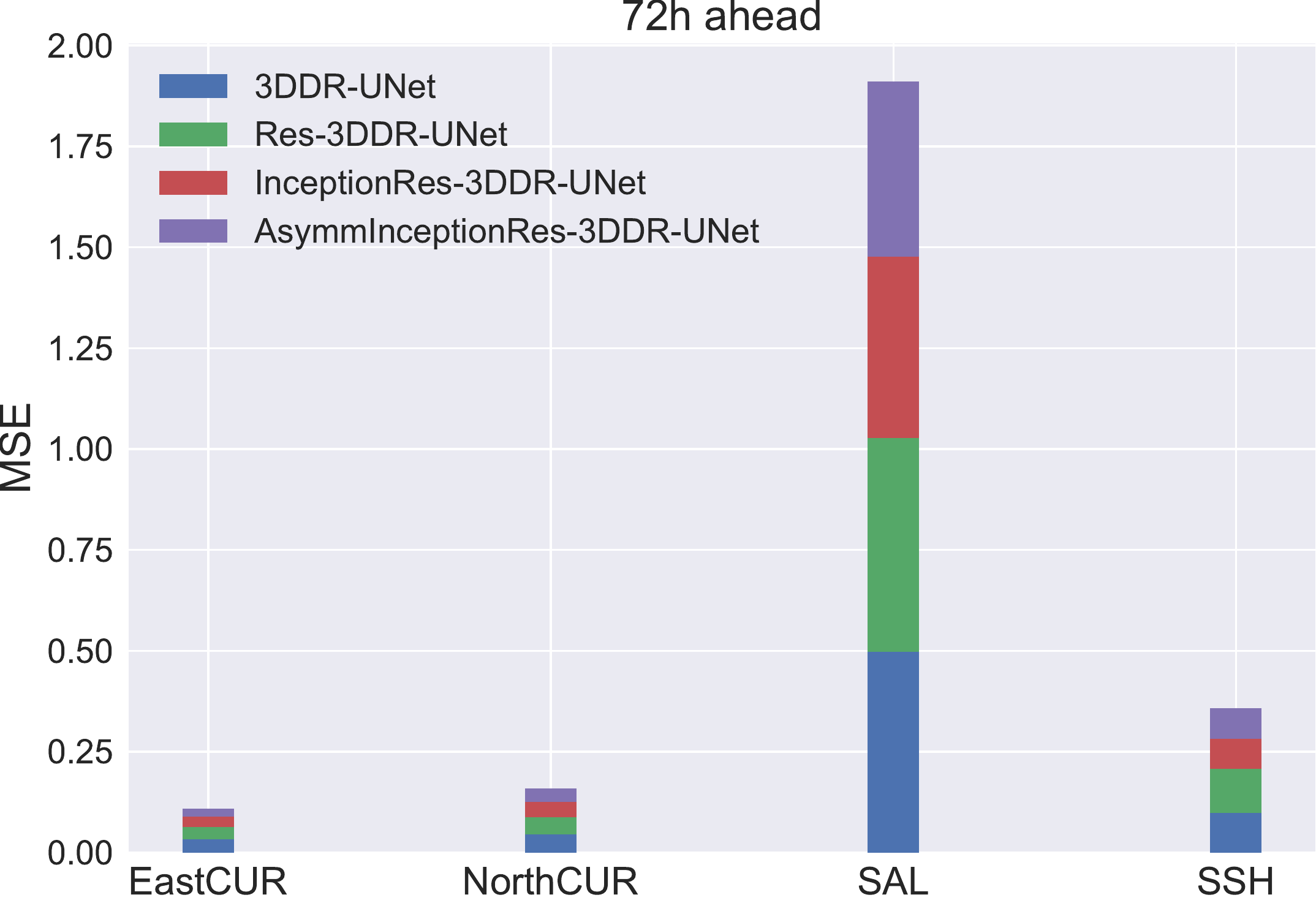}}
    \\\vskip 0.5pt plus 0.25fil
    \caption{MSE of individual sea elements for the different models. The predictions are performed using (a) 12h, (b) 24h, (c) 48h, and (d) 72h ahead.}
    \label{fig:mseVariables2}
\end{figure*}

This section presents the results obtained from the described experiments. The obtained MSEs of all the models for each of the configurations are tabulated in Table \ref{tab:results}. These results correspond to testing of the models on the four combined seasons. It can be observed that for all the models the prediction error increases as the number of hours ahead increases. AsymmInceptionRes-3DDR-UNet model performs better than the other models in almost all the scenarios. This improvement in the performance is more apparent as the number of hours ahead increases.  
\begin{table}[!htbp]
    \centering
    \scriptsize{
    \renewcommand{\arraystretch}{1.5}
    \caption{Test MSE of all models in all four seasons.}
    \label{tab:results}
    \begin{tabular}{l l l l l}
    \Xhline{3\arrayrulewidth}
    &\multicolumn{4}{c}{\textbf{Hours ahead}} \\
    \Xhline{3\arrayrulewidth}
    \multirow{2}{*}{\textbf{Model}}& 
    \multirow{2}{*}{\textbf{12h}} & 
    \multirow{2}{*}{\textbf{24h}} & 
    \multirow{2}{*}{\textbf{48h}} & 
    \multirow{2}{*}{\textbf{72h}}\\
     & & & & \\\Xhline{3\arrayrulewidth}
         \textbf{3DDR-UNet} & 5.40e-02	 & 7.78e-02	 & 1.21e-01	& 1.69e-01	\\ 
         \textbf{Res-3DDR-UNet} & 6.34e-02	& 7.99e-02	& 1.42e-01	& 1.77e-01	\\ 
         \textbf{InceptionRes-3DDR-UNet} & 5.19e-02	 & \underline{7.08e-02}	& 1.20e-01	 & 1.47e-01	 \\ 
         \textbf{AsymmInceptionRes-3DDR-UNet} & \underline{5.15e-02} & 7.56e-02	& \underline{1.17e-01}	& \underline{1.41e-01}	\\ 
        \Xhline{3\arrayrulewidth}
    \end{tabular}
    }
\end{table}
In Table \ref{tab:resultsSeasons}, the test MSE of each model is displayed separated for each season. Similar to the previous results, AsymmInceptionRes-3DDR-UNet outperforms the other discussed models in most the setups. One may also observe that seasons with more changing weather conditions such as winter makes it more challenging for the networks to learn the underlying complex patterns. All of the four models yield noticeably better predictions in seasons with more stable weather, like summer. Furthermore, a comparison of the final number of convolutional layers of each model can be found in Fig. \ref{fig:numberParams}. Fig. \ref{fig:mseVariables} shows the obtained MSE of the test data for each sea element. Fig. \ref{fig:mseVariables2} (a,b,c,d) corresponds to 12, 24, 48 and 72 hours ahead prediction respectively. In general, we observe that seawater salinity is the most challenging variables to be predicted among the considered variables in this study.

\begin{table*}[!htbp]
    \centering
 \scriptsize{
    \renewcommand{\arraystretch}{1.5}
    \caption{Test MSE of all models in different seasons.}
    \label{tab:resultsSeasons}
    \begin{tabular}{l l l l l l}
    \Xhline{3\arrayrulewidth}
    \multirow{2}{*}{\textbf{Season}}& 
    \multirow{2}{*}{\textbf{Model}}& 
    \multirow{2}{*}{\textbf{12h ahead}} & 
    \multirow{2}{*}{\textbf{24h ahead}} & 
    \multirow{2}{*}{\textbf{48h ahead}} & 
    \multirow{2}{*}{\textbf{72h ahead}}\\
    & & & & \\\Xhline{3\arrayrulewidth}
        \multirow{4}{*}{\textbf{Spring}}
         & \textbf{3DDR-UNet} & 7.45e-02 & 9.95e-02 & 1.37e-01 & 2.14e-01\\ 
         & \textbf{Res-3DDR-UNet} & 8.37e-02 & 1.01e-01 & 1.65e-01 & 1.69e-01 \\ 
         & \textbf{InceptionRes-3DDR-UNet} & \underline{6.63e-02} & \underline{9.37e-02} & 1.43e-01 & 1.78e-01 \\ 
         & \textbf{AsymmInceptionRes-3DDR-UNet} & 6.98e-02 & 9.56e-02 & \underline{1.38e-01} & \underline{1.66e-01} \\ 
    \Xhline{3\arrayrulewidth}
        \multirow{4}{*}{\textbf{Summer}}
         & \textbf{3DDR-UNet} & 4.10e-02 & 7.92e-02 & 1.37e-01 & 1.69e-01 \\ 
         & \textbf{Res-3DDR-UNet} & 5.23e-02 & 6.85e-02 & 1.23e-01 & 1.97e-01 \\ 
         & \textbf{InceptionRes-3DDR-UNet} & 4.37e-02 & \underline{6.55e-02} & 1.06e-01 & 1.49e-01  \\ 
         & \textbf{AsymmInceptionRes-3DDR-UNet} & \underline{4.03e-02} & 7.88e-02 & \underline{1.04e-01} & \underline{1.43e-01}  \\  
    \Xhline{3\arrayrulewidth}
        \multirow{4}{*}{\textbf{Autumn}}
         & \textbf{3DDR-UNet} & 3.36e-02 & 4.81e-02 & 6.95e-02 & 7.89e-02 \\ 
         & \textbf{Res-3DDR-UNet} & 3.97e-02 & 5.61e-02 & 8.92e-02 & 1.27e-01 \\ 
         & \textbf{InceptionRes-3DDR-UNet} & 3.29e-02 & \underline{4.31e-02} & 6.98e-02 & \underline{6.75e-02} \\ 
         & \textbf{AsymmInceptionRes-3DDR-UNet} & \underline{2.97e-02} & 4.66e-02 & \underline{6.79e-02} & 7.11e-02 \\ 
    \Xhline{3\arrayrulewidth}
        \multirow{4}{*}{\textbf{Winter}}
         & \textbf{3DDR-UNet} & 6.62e-02 & 8.51e-02 & 1.62e-01 & 1.90e-01 \\ 
         & \textbf{Res-3DDR-UNet} & 7.66e-02 & 9.44e-02 & 1.88e-01 & 1.99e-01 \\ 
         & \textbf{InceptionRes-3DDR-UNet} & \underline{6.41e-02} & \underline{8.18e-02} & 1.63e-01 & 1.71e-01 \\ 
         & \textbf{AsymmInceptionRes-3DDR-UNet} & 6.51e-02 & 8.28e-02 & \underline{1.58e-01} & \underline{1.62e-01} \\ 
    \Xhline{3\arrayrulewidth}
    \end{tabular}
    }
\end{table*}

\begin{figure}[!htbp]
    \centering
    \subfloat[]{\includegraphics[width=1\linewidth]{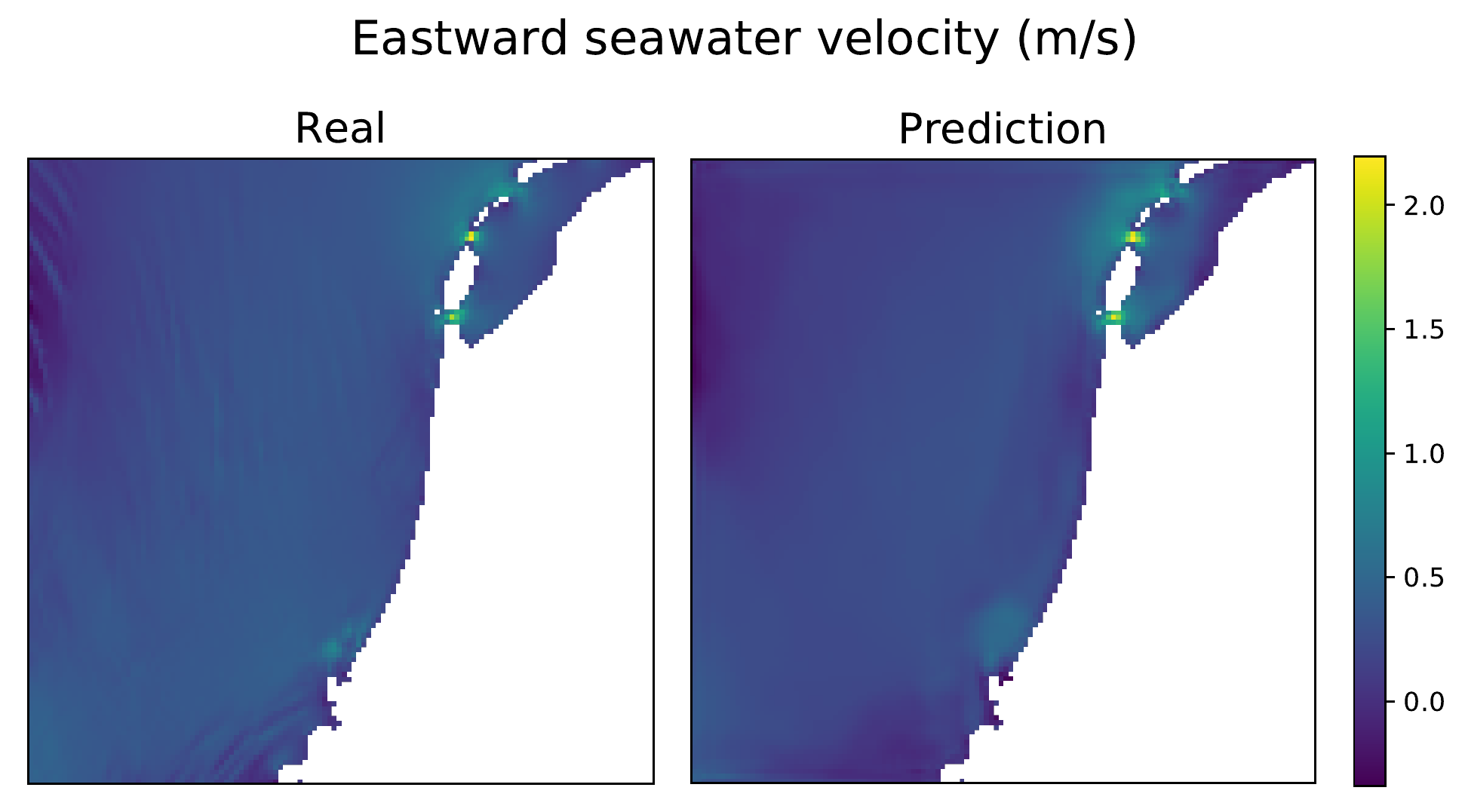}} \\
    \subfloat[]{\includegraphics[width=1.014\linewidth]{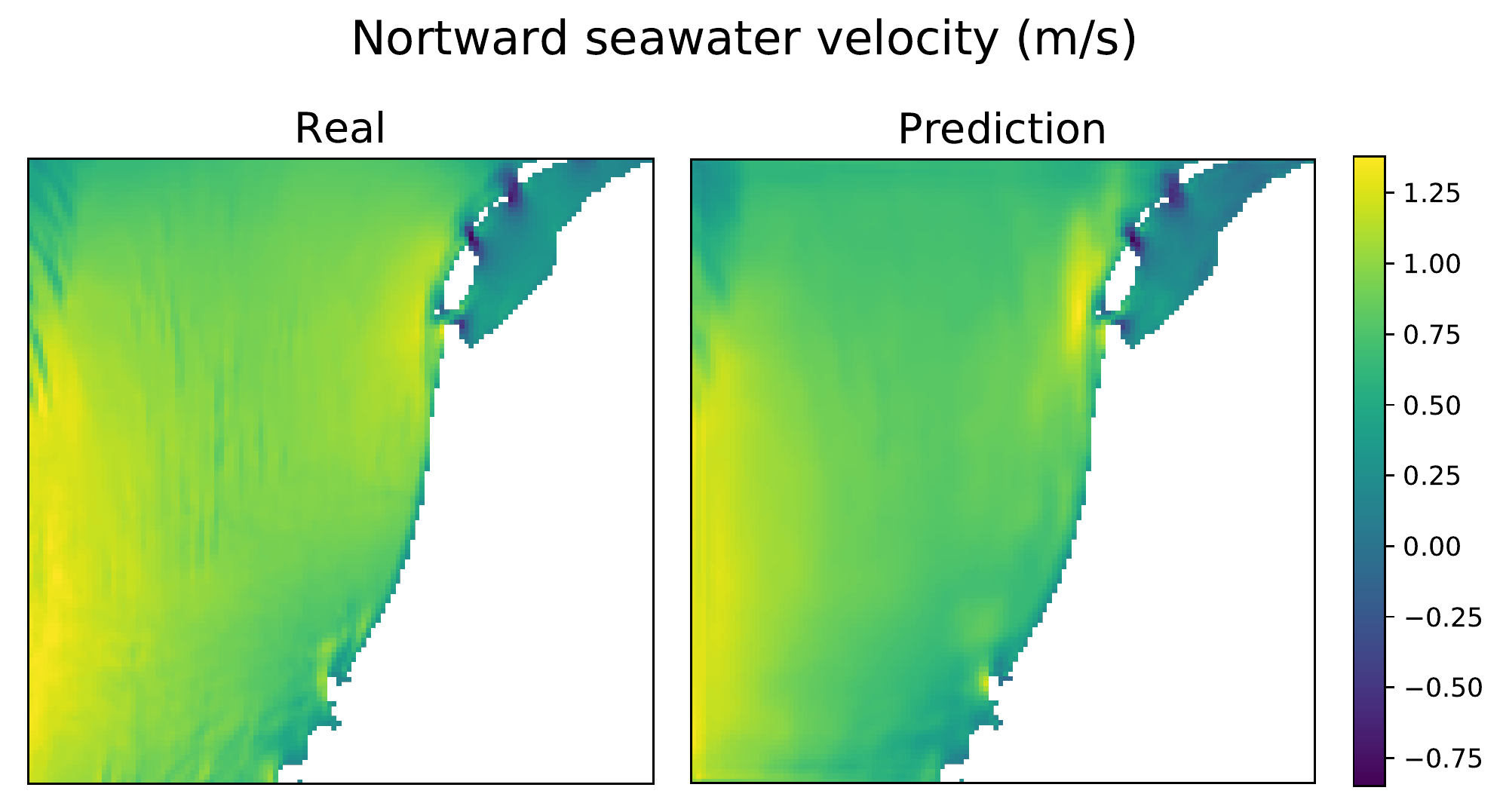}} \\
    \hspace{-0.22cm}
    \subfloat[]{\includegraphics[width=0.992\linewidth]{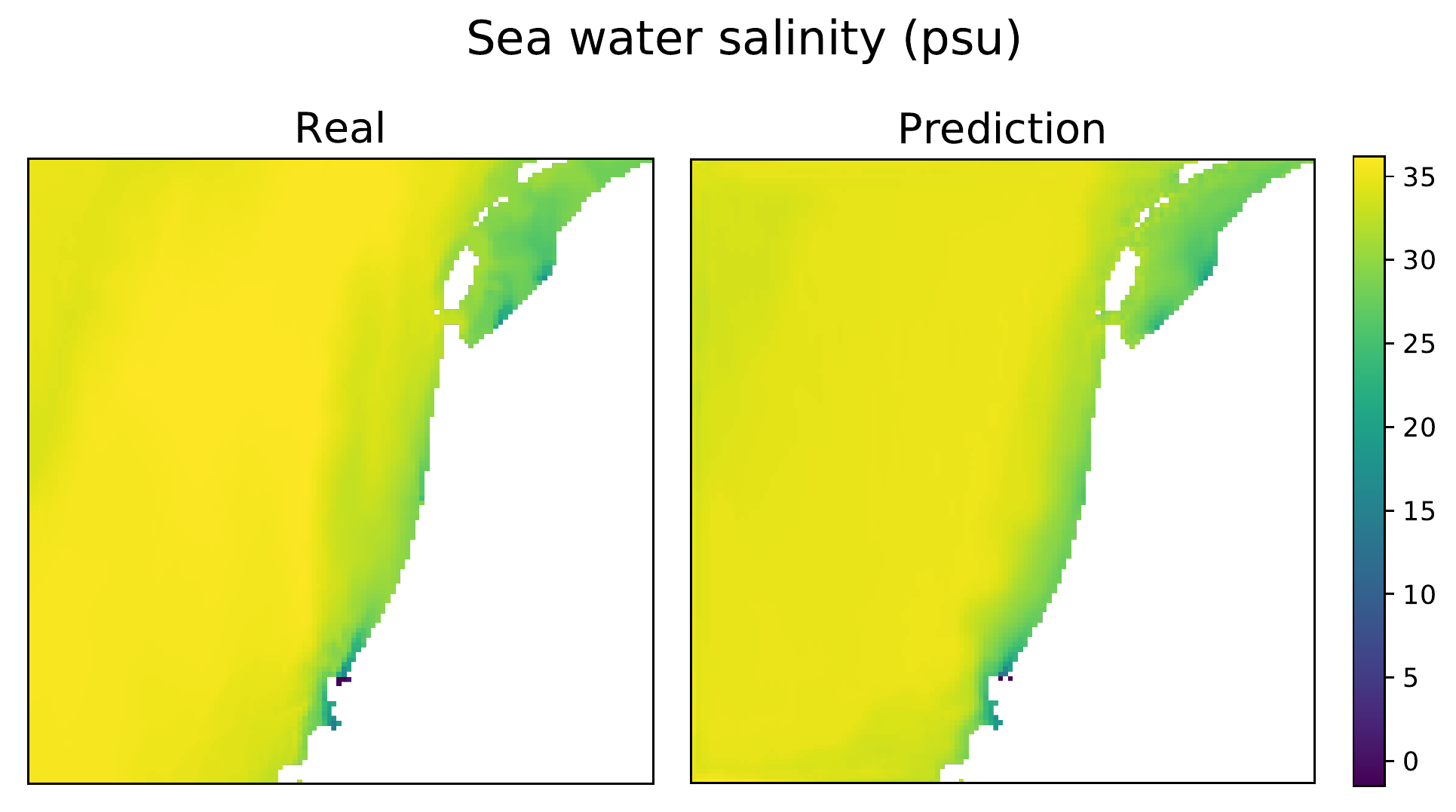}} \\
    \hspace{-0.1729cm}
    \subfloat[]{\includegraphics[width=1.011985\linewidth]{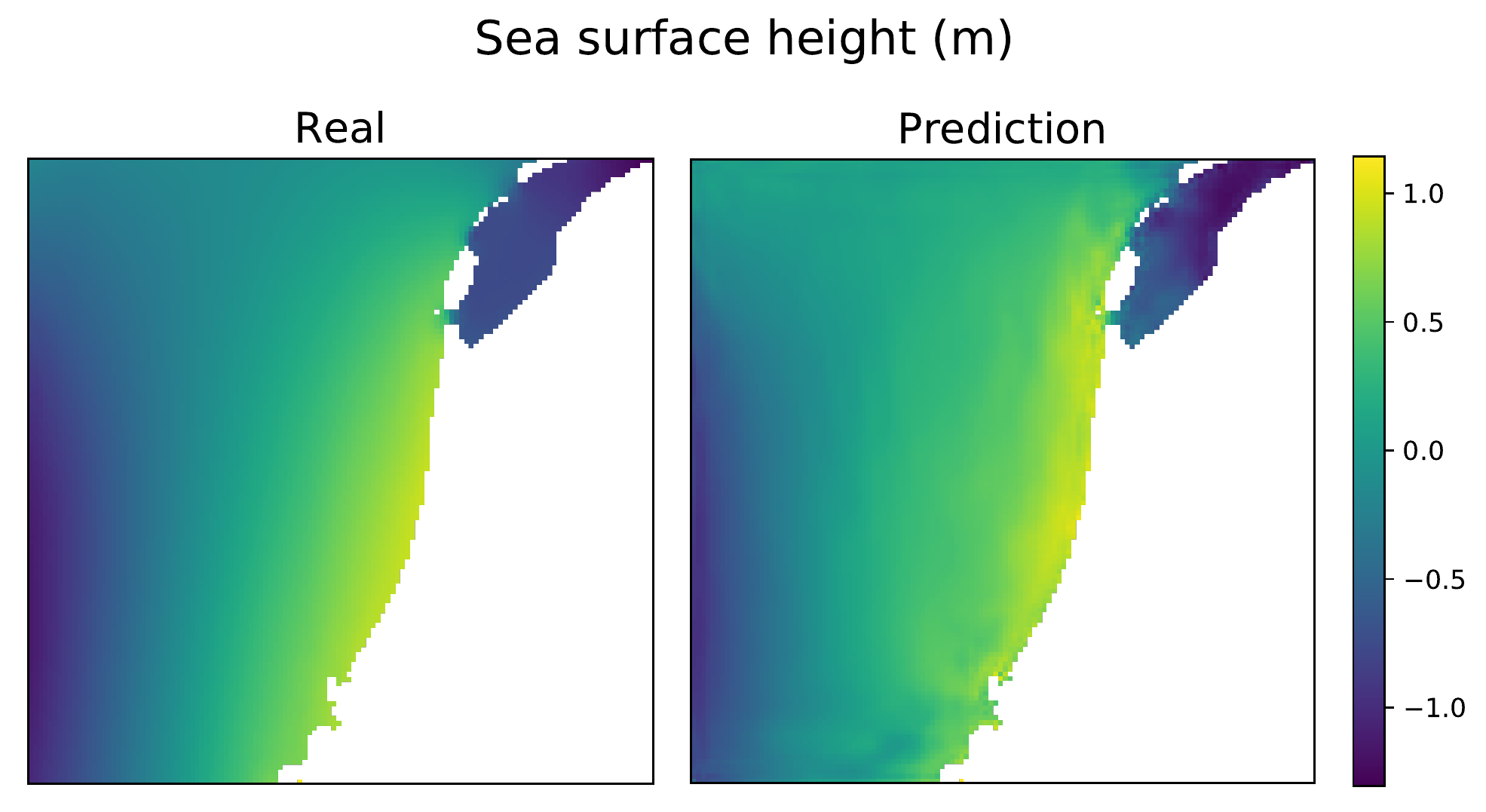}}
    \caption{AsymmInceptionRes-3DDR-UNet's prediction of all variables 48 hours ahead in winter. The data is scaled back to its original values.}
    \label{fig:ExamplePredictionWinter}
\end{figure}

An example of the 48h ahead forecast with AsymmInceptionRes-3DDR-UNet model during winter is shown in Fig. \ref{fig:ExamplePredictionWinter}. As it can be seen, the forecast is considerably accurate, even when it comes to seawater velocity (EastCUR and NorthCUR), which contains quite different areas. The obtained results suggest that inclusion of parallel branches of convolutions, presented in InceptionRes-3DDR-UNet and Asymm InceptionRes-3DDR-UNet models has led to a more noticeable performance improvement.

\section{Conclusion}\label{sec:conclusion}
In this paper, four new models based on the U-Net architecture are introduced for multi-step ahead coastal sea elements prediction. The proposed models are examined under different setups, i.e. different seasons and numbers of hours ahead.
Among the discussed models, AsymmInceptionRes-3DDR-UNet and InceptionRes-3DDR-UNet have shown superior performance thanks to the use of parallel convolutions. However, the incorporation of asymmetric convolutions and additional parallel branches make the AsymmInceptionRes-3DDR-UNet perform slightly better than the latter, yielding the most promising results on the studied tasks. 
The scripts and models used in this paper can be found in \url{https://github.com/jesusgf96/Sea-Elements-Prediction-UNet-Based-Models}.

\section*{Acknowledgment}
Simulations were performed with computing resources granted by RWTH Aachen University.


\bibliography{Main}

\end{document}